\documentstyle[aclap]{article}

\title{FAST PARSING USING PRUNING AND GRAMMAR SPECIALIZATION}

\author{Manny Rayner and David Carter \\
SRI International \\ 
Suite 23, Millers Yard\\
Cambridge CB2 1RQ\\
United Kingdom \\
{\tt manny@cam.sri.com, dmc@cam.sri.com}}

\begin{document}

\maketitle
\vspace{-0.5in}

\begin{abstract}

We show how a general grammar may be automatically adapted for fast
parsing of utterances from a specific domain by means of constituent
pruning and grammar specialization based on explanation-based
learning. These methods together give an order of magnitude increase
in speed, and the coverage loss entailed by grammar specialization is
reduced to approximately half that reported in previous
work. Experiments described here suggest that the loss of coverage has
been reduced to the point where it no longer causes significant
performance degradation in the context of a real application.

\end{abstract}

\bibliographystyle{acl}

\section{Introduction}
\label{Introduction}

Suppose that we have a general grammar for English, or some other
natural language; by this, we mean a grammar which encodes most of the
important constructions in the language, and which is intended to be
applicable to a large range of different domains and applications. The
basic question attacked in this paper is the following one: can such a
grammar be concretely useful if we want to process input from a {\it
specific} domain? In particular, how can a parser that uses a general
grammar achieve a level of efficiency that is practically acceptable?

The central problem is simple to state. By the very nature of its
construction, a general grammar allows a great many theoretically
valid analyses of almost any non-trivial sentence.  However, in the
context of a specific domain, most of these will be extremely
implausible, and can in practice be ignored. If we want efficient
parsing, we want to be able to focus our search on only a small
portion of the space of theoretically valid grammatical analyses.

One possible solution is of course to dispense with the idea of using
a general grammar, and simply code a new grammar for each domain.
Many people do this, but one cannot help feeling that something is
being missed; intuitively, there are many domain-independent
grammatical constraints, which one would prefer only to need to code
once. In the last ten years, there have been a number of attempts to
find ways to automatically adapt a general grammar and/or parser to
the sub-language defined by a suitable training corpus. For
example, \cite{BriscoeCarroll:CL} train an LR parser based on a
general grammar to be able to distinguish between likely and unlikely
sequences of parsing actions; \cite{MenloSorts} automatically infer
sortal constraints, that can be used to rule out otherwise grammatical
constituents; and \cite{GrammarPruning} describes methods that reduce
the size of a general grammar to include only rules actually useful
for parsing the training corpus.

The work reported here is a logical continuation of two specific
strands of research aimed in this general direction. The first is the
popular idea of {\it statistical tagging} e.g.\ 
\cite{DeRose,XeroxTagger,ChurchTagger}.  Here, the basic idea is that
a given small segment $S$ of the input string may have several
possible analyses; in particular, if $S$ is a single word, it may
potentially be any one of several parts of speech.  However, if a
substantial training corpus is available to provide reasonable
estimates of the relevant parameters, the immediate context
surrounding $S$ will usually make most of the locally possible
analyses of $S$ extremely implausible.  In the specific case of
part-of-speech tagging, it is well-known \cite{DeMarckenACL} that a
large proportion of the incorrect tags can be eliminated ``safely'',
i.e.\ with very low risk of eliminating correct tags. In the present
paper, the statistical tagging idea is generalized to a method called
``constituent pruning''; this acts on local analyses of phrases
normally larger than single-word units.

Constituent pruning is a bottom-up approach, and is complemented 
by a second, top-down, method based on {\it Explanation-Based
Learning} (EBL; \cite{MitchellEBL,HarmelenBundyEBL}). This part of the
paper is essentially an extension and generalization of the line of
work described in
\cite{FGCS-88,EBL-ARPA,EBL-IJCAI-91,SLT-report-EBL,Christer-ACL-94}. Here,
the basic idea is that grammar rules tend in any specific domain to
combine much more frequently in some ways than in others. Given a
sufficiently large corpus parsed by the original, general, grammar, it
is possible to identify the common combinations of grammar rules and
``chunk'' them into ``macro-rules''. The result is a
``specialized'' grammar; this has a larger number of rules, but a
simpler structure, allowing it in practice to be parsed very much more
quickly using an LR-based method \cite{Christer-COLING-94}. The coverage of
the specialized grammar is a strict subset of that of the original
grammar; thus any analysis produced by the specialized grammar is
guaranteed to be valid in the original one as well. The practical
utility of the specialized grammar is largely determined by the loss
of coverage incurred by the specialization process.

The two methods, constituent pruning and grammar specialization, are
combined as follows. The rules in the original, general,
grammar are divided into two sets, called {\it phrasal} and {\it
non-phrasal} respectively. Phrasal rules, the majority of which
define non-recursive noun phrase constructions, are used
as they are; non-phrasal rules are combined using EBL into chunks,
forming a specialized grammar which is then compiled further into a
set of LR-tables. Parsing proceeds by interleaving constituent
creation and deletion. First, the lexicon and morphology rules are
used to hypothesize word analyses.  Constituent pruning then removes
all sufficiently unlikely edges.  Next, the phrasal rules are applied
bottom-up, to find all possible phrasal edges, after which unlikely
edges are again pruned. Finally, the specialized grammar is used to
search for full parses. The scheme is fully implemented within a
version of the Spoken Language Translator system
\cite{SLT-HLT,SLT-report}, and is normally applied to input in the
form of small lattices of hypotheses produced by a speech recognizer.

The rest of the paper is structured as follows. Section~\ref{Pruning}
describes the constituent pruning method. Section~\ref{EBL} describes
the grammar specialization method, focusing on how the current work
extends and improves on previous results. Section~\ref{Experiments}
describes experiments where the constituent pruning/grammar
specialization method was used on sets of previously unseen speech
data.  Section~\ref{Section:Conclusions} concludes and sketches further
directions for research, which we are presently in the process of
investigating.

\section{Constituent Pruning}
\label{Pruning}

Before both the phrasal and full parsing stages, the constituent table
(henceforth, the chart) is pruned to remove edges that are relatively
unlikely to contribute to correct analyses.

For example, after the string ``Show flight D L three one two'' is
lexically analysed, edges for ``D'' and ``L'' as individual characters
are pruned because another edge, derived from a lexical entry for ``D
L'' as an airline code, is deemed far more plausible. Similarly, edges
for ``one'' as a determiner and as a noun are pruned because, when
flanked by two other numbers, ``one'' is far more likely to function
as a number.

Phrasal parsing then creates a number of new edges, including one for
``flight D L three one two'' as a noun phrase. This edge is deemed far
more likely to serve as the basis for a correct full parse than any of
the edges spanning substrings of this phrase; those edges, too, are
therefore pruned.  As a result, full parsing is very quick, and only one
analysis (the correct one) is produced for the sentence. In the
absence of pruning, processing takes over eight times as long and
produces 37 analyses in total.

\subsection{The pruning algorithm}

Our algorithm estimates the probability of correctness of each edge:
that is, the probability that the edge will contribute to the correct
full analysis of the sentence (assuming there is one), given certain
lexical and/or syntactic information about it. Values on each
criterion (selection of pieces of information) are derived from
training corpora by maximum likelihood estimation followed by
smoothing. That is, our estimate for the probability that an edge with
property $P$ is correct is (modulo smoothing) simply the number of
times edges with property $P$ occur in {\it correct} analyses in
training divided by the number of times such edges are {\it created}
during the analysis process in training.

The current criteria are:

\begin{itemize}

\item The {\it left bigram} score: the probability of correctness of
an edge considering only the following data about it:
\begin{itemize}
\item its {\it tag} (corresponding to its major category symbol plus, for
a few categories, some additional distinctions derived from feature
values);
\item for a lexical edge, its {\it word} or {\it semantic word class}
(words with similar distributions, such as city names, are grouped
into classes to overcome data sparseness); or for a phrasal edge, the
name of the final (topmost) {\it grammar rule} that was used to create
it;
\item the tag of a neighbouring edge immediately to its left. If there
are several left neighbours, the one giving the highest probability is
used.
\end{itemize}

\item The {\it right bigram} score: as above, but considering right
neighbours.

\item The {\it unigram} score: the probability of correctness of an
edge considering only the tree of grammar rules,
with words or word classes at the leaves, that gave rise to it.
For a lexical edge, this reduces to its word or word class, and its tag.

\end{itemize}

Other criteria, such as trigrams and finer-grained tags, are obviously
worth investigating, and could be applied straightforwardly within the
framework described here.

The minimum score derived from any of the criteria applied is deemed
initially to be the score of the constituent. That is, an assumption
of full statistical dependence \cite{Yarowsky}, rather than the more
common full independence, is made.\footnote{If events $E_1, E_2, ...,
E_n$ are fully independent, then the joint probability $P(E_1 \wedge
... \wedge E_n)$ is the product of $P(E_1) ... P(E_n)$, but if they
are maximally dependent, it is the minimum of these values. Of course,
neither assumption is any more than an approximation to the truth; but
assuming dependence has the advantage that the estimate of the joint
probability depends much less strongly on $n$, and so estimates for
alternative joint events can be directly compared, without any
possibly tricky normalization, even if they are composed of different
numbers of atomic events. This property is desirable: different
(sub-)paths through a chart may span different numbers of edges, and
one can imagine evaluation criteria which are only defined for some
kinds of edge, or which often duplicate information supplied by other
criteria. Taking minima means that the pruning of an edge results from
it scoring poorly on one criterion, regardless of other, possibly good
scores assigned to it by other criteria. This fits in with the fact
that on the basis of local information alone it is not usually
possibly to predict with confidence that a particular edge is highly
{\it likely} to contribute to the correct analysis (since global
factors will also be important) but it often is possible to spot
highly {\it unlikely} edges. In other words, our training procedure
yields far more probability estimates close to zero than close to
one.}  When recognizer output is being processed, however, the
estimate from each criterion is in fact multiplied by a further
estimate derived from the acoustic score of the edge: that is, the
score assigned by the speech recognizer to the best-scoring sentence
hypothesis containing the word or word string for the edge in
question. Multiplication is used here because acoustic and
lexicosyntactic likelihoods for a word or constituent would appear to
be more nearly fully independent than fully dependent, being based on
very different kinds of information.

Next, account is taken of the connectivity of the chart. Each {\it
vertex} of the chart is labelled with the score of the best path
through the chart that visits that vertex. In accordance with the
dependence assumption, the score of a path is defined as the minimum
of the scores of its component edges. Then the score of each edge is
recalculated to be the minimum of its existing score and the scores of
its start and end vertices, on the grounds that a constituent, however
intrinsically plausible, is not worth preserving if it does not occur
on any plausible paths.

Finally, a pruning threshold is calculated as the score of the best
path through the chart multiplied by a certain fraction.  For the
first pruning phase we use 1/20, and for the second, 1/150, although
performance is not very sensitive to this. Any constituents scoring
less than the threshold are pruned out.

\subsection{Relation to other pruning methods}

As the example above suggests, judicious pruning of the chart at
appropriate points can greatly restrict the search space and speed up
processing. Our method has points of similarity with some very recent
work in Constraint Grammar\footnote{Christer Samuelsson, personal
communication, 8th April 1996; see \cite{CG} for background.} and is an
alternative to several other, related schemes.

Firstly, a remarked earlier, it generalizes {\it tagging}: it not only
adjudicates between possible labels for the same word, but can also
use the existence of a constituent over one span of the chart as
justification for pruning another constituent over another span,
normally a subsumed one, as in the ``D L'' example. This is especially
true in the second stage of pruning, when many constituents of
different lengths have been created. Furthermore, it applies equally
well to lattices, rather than strings, of words, and can take account
of acoustic plausibility as well as syntactic considerations.

Secondly, our method is related to {\it beam search} \cite{Woods}.
In beam search, incomplete parses of an utterance are pruned or
discarded when, on some criterion, they are significantly less
plausible than other, competing parses. This pruning is fully
interleaved with the parsing process. In contrast, our pruning takes
place only at certain points: currently before parsing begins, and
between the phrasal and full parsing stages. Potentially, as with any
generate-and-test algorithm, this can mean efficiency is reduced: some
paths will be explored that could in principle be pruned
earlier. However, as the results in section \ref{Experiments} below
will show, this is not in practice a serious problem, because the
second pruning phase greatly reduces the search space in preparation
for the potentially inefficient full parsing phase. Our method has the
advantage, compared to beam search, that there is no need for any
particular search order to be followed; when pruning takes place, all
constituents that could have been found at the stage in question are
guaranteed already to exist.

Thirdly, our method is a generalization of the strategy employed by
\cite{McCord}. McCord interleaved parsing with pruning in the same way
as us, but only compared constituents over the same span and with the
same major category. Our comparisons are more global and therefore can
result in more effective pruning.

\section{Grammar specialization}
\label{EBL}
\label{Section:EBL}

As described in Section~\ref{Introduction} above, the non-phrasal
grammar rules are subjected to two phases of processing. In the first,
``EBL learning'' phase, a parsed training corpus is used to identify
``chunks'' of rules, which are combined by the EBL algorithm into
single macro-rules. In the second phase, the resulting set of
``chunked'' rules is converted into LR table form, using the
method of \cite{Christer-COLING-94}.

There are two main parameters that can be adjusted in the EBL learning
phase. Most simply, there is the size of the training corpus; 
a larger training corpus means a smaller loss of coverage
due to grammar specialization. (Recall that grammar specialization
in general trades coverage for speed). Secondly, there is the question
of how to select the rule-chunks that will be turned into macro-rules.
At one limit, the whole parse-tree for each training example is turned
into a single rule, resulting in a specialized grammar all of whose
derivations are completely ``flat''. These grammars can be parsed
extremely quickly, but the coverage loss is in practice unacceptably
high, even with very large training corpora. At the opposite extreme,
each rule-chunk consists of a single rule-application; this yields
a specialized grammar identical to the original one. The challenge is 
to find an intermediate solution, which specializes the grammar
non-trivially without losing too much coverage.

Several attempts to find good ``chunking criteria'' are described in
the papers by Rayner and Samuelsson quoted above. In
\cite{SLT-report-EBL}, a simple scheme is given, which creates rules
corresponding to four possible units: full utterances, recursive NPs,
PPs, and non-recursive NPs. A more elaborate scheme is given in
\cite{Christer-ACL-94}, where the ``chunking criteria'' are learned
automatically by an entropy-minimization method; the results, however,
do not appear to improve on the earlier ones. In both cases, the
coverage loss due to grammar specialization was about 10 to
12\% using training corpora with about 5,000 examples. In
practice, this is still unacceptably high for most applications.

Our current scheme is an extension of the one from
\cite{SLT-report-EBL}, where the rule-chunks are trees of non-phrasal
rules whose roots and leaves are categories of the following possible
types: full utterances, utterance units, imperative VPs, NPs, relative
clauses, VP modifiers and PPs. The resulting specialized grammars are
forced to be non-recursive, with derivations being a maximum of six
levels deep.  This is enforced by imposing the following dominance
hierarchy between the possible categories:

\begin{center}
{\tt 
utterance $>$ utterance\_unit $>$ imperative\_VP \\
$>$ NP $>$ \{rel, VP\_modifier\} $>$ PP
}
\end{center}

\noindent
The precise definition of the rule-chunking criteria is quite simple, and
is reproduced in the appendix.

Note that only the non-phrasal rules are used as input to the chunks
from which the specialized grammar rules are constructed. This has two
important advantages. Firstly, since all the phrasal rules are
excluded from the specialization process, the coverage loss associated
with missing combinations of phrasal rules is eliminated. As the
experiments in the next section show, the resulting improvement is
quite substantial. Secondly, and possibly even more importantly, the
number of specialized rules produced by a given training corpus is
approximately halved. The most immediate consequence is that much
larger training corpora can be used before the specialized grammars
produced become too large to be handled by the LR table compiler. If
both phrasal and non-phrasal rules are used, we have been unable to
compile tables for rules derived from training sets of over 6,000
examples (the process was killed after running for about six hours on
a Sun Sparc 20/HS21, SpecINT92=131.2). Using only non-phrasal rules,
compilation of the tables for a 15,000 example training set required
less than two CPU-hours on the same machine.

\section{Experiments}
\label{Experiments}

This section describes a number of experiments carried out to 
test the utility of the theoretical ideas presented above.
The basic corpus used was a set of 16,000 utterances from the
Air Travel Planning (ATIS; \cite{ATIS}) domain. 
All of these utterances were available in text form; 15,000
of them were used for training, with 1,000 held out for test 
purposes. Care was taken to ensure not just that the utterances
themselves, but also the {\it speakers} of the utterances were
disjoint between test and training data; as pointed out in
\cite{ARPA-NBEST}, failure to observe these precautions can 
result in substantial spurious improvements in test data results.

The 16,000 sentence corpus was analysed by the SRI Core Language
Engine \cite{CLE}, using a lexicon extended to cover the ATIS domain
\cite{SLT-report-grammar}. All possible grammatical analyses of each
utterance were recorded, and an interactive tool was used to allow a
human judge to identify the correct and incorrect readings of each
utterance. The judge was a first-year undergraduate student with a
good knowledge of linguistics but no prior experience with the system;
the process of judging the corpus took about two and a half
person-months.
The input to the EBL-based grammar-specialization
process was limited to readings of corpus utterances that had been
judged correct.  When utterances had more than one correct reading, a
preference heuristic was used to select the most plausible one.

Two sets of experiments were performed. In the first, increasingly
large portions of the training set were used to train specialized
grammars. The coverage loss due to grammar specialization was then
measured on the 1,000 utterance test set. The experiment was carried
out using both the chunking criteria from \cite{SLT-report-EBL} (the
``Old'' scheme), and the chunking criteria described in
Section~\ref{Section:EBL} above (the ``New'' scheme). The results
are presented in Table~1.

\begin{table}
\begin{center}
\begin{tabular}{|r|r|r|r|r|} \hline
Examples & \multicolumn{2}{c}{Old scheme} & \multicolumn{2}{c|}{New
scheme} \\ \hline
           &   Rules  &  Loss        & Rules      &  Loss \\ \hline
100      &       100  &    47.8\%    &      69    &     35.5\%  \\
250      &       181  &    37.6\%    &     126    &     21.8\%  \\
500      &       281  &    27.6\%    &     180    &     14.7\%  \\
1000     &       432  &    22.7\%    &     249    &     10.8\%  \\
3000     &       839  &    14.9\%    &     455    &      7.8\%  \\
5000     &      1101  &    11.2\%    &     585    &      6.6\%  \\
7000     &      1292  &    10.4\%    &     668    &      6.0\%  \\
11000    &      1550  &     9.8\%    &     808    &      5.8\%  \\
15000    &      1819  &     8.7\%    &     937    &      5.0\%  \\ \hline
\end{tabular}
\caption{EBL rules and EBL coverage loss against number of training
examples}
\end{center}
\label{Table:EBL-coverage}
\end{table}

The second set of experiments tested more directly the effect of
constituent pruning and grammar specialization on the Spoken Language
Translator's speed and coverage; in particular, coverage was measured
on the real task of translating English into Swedish, rather than the
artificial one of producing a correct QLF analysis. To this end, the
first 500 test-set utterances were presented in the form of speech
hypothesis lattices derived by aligning and conflating the top five
sentence strings produced by a version of the DECIPHER (TM) recognizer
\cite{Murveit:93}.  The lattices were analysed by four different
versions of the parser, exploring the different combinations of
turning constituent pruning on or off, and specialized versus
unspecialized grammars. The specialized grammar used the ``New''
scheme, and had been trained on the full training set. Utterances
which took more than 90 CPU seconds to process were timed out and
counted as failures.


The four sets of outputs from the parser were then translated into
Swedish by the SLT transfer and generation mechanism
\cite{SLT-report}. Finally, the four sets of candidate translations
were pairwise compared in the cases where differing translations had
been produced. We have found this to be an effective way of evaluating
system performance. Although people differ widely in their judgements
of whether a given translation can be regarded as ``acceptable'', it
is in most cases surprisingly easy to say which of two possible
translations is preferable. The last two tables summarize the
results. Table~2 
gives the average
processing times per input lattice for each type of processing (times
measured running SICStus Prolog 3\#3 on a SUN Sparc 20/HS21),
showing how the time is divided between the various processing
phases.  Table~3 
shows the relative scores of the
four parsing variants, measured according to the ``preferable
translation'' criterion.

\section{Conclusions and further directions}
\label{Section:Conclusions}

Table~2 
indicates that EBL and pruning each make
processing about three times faster; the combination of both gives a
factor of about nine.  In fact, as the detailed breakdown
shows, even this underestimates the effect on the main parsing phase:
when both pruning and EBL are operating, processing times for other
components (morphology, pruning and preferences) become the dominant
ones. As we have so far expended little effort on optimizing these
phases of processing, it is reasonable to expect substantial further
gains to be possible.

\begin{table}
\begin{center}
\begin{tabular}{|r|r|r|r|r|} \hline
                 & E--   & E+    & E--   & E+  \\
                 & P--   & P--   & P+    & P+ \\ \hline
Morph/lex lookup &  0.53 & 0.54  & 0.54  & 0.49     \\ \hline
Phrasal parsing  &  0.27 & 0.28  & 0.14  & 0.14     \\ \hline
Pruning          & --    & --    & 0.57  & 0.56       \\ \hline
Full parsing     & 12.42 & 2.61  & 3.04  & 0.26       \\ \hline
Preferences      &  3.63 & 1.57  & 1.27  & 0.41       \\ \hline \hline
TOTAL            & 16.85 & 5.00  & 5.57  & 1.86       \\ \hline
\end{tabular}
\caption{Breakdown of average time spent on each processing phase
for each type of processing (seconds per utterance)}
\end{center}
\label{Table:FourWayTab}
\end{table}

\begin{table}
\begin{center}
\begin{tabular}{|c|c|c|c|c|} \hline
         & E--   & E+    & E--   & E+         \\
         & P--   & P--   & P+    & P+         \\ \hline
E--/P--  &       &12--24 &25--63 & 24--65     \\ \hline
E+/P--   &24--12 &       &31--50 & 26--47     \\ \hline
E--/P+   &63--25 &50--31 &       & 5--8       \\ \hline
E+/P+    &65--24 &47--26 &8--5   &            \\ \hline
\end{tabular}
\caption{Comparison between translation results on the four different
analysis alternatives, measured on the 500-utterance test set. The
entry for a given row and column holds two figures, showing
respectively the number of examples where the ``row'' variant produced
a better translation than the ``column'' variant and the number where
it produced a worse one. Thus for example ``EBL+/pruning+'' was better
than ``EBL--/pruning--'' on 65 examples, and worse on 24.}
\end{center}
\label{Table:TimeBreakdown}
\end{table}

Even more interestingly, Table~3
shows that real system performance, in terms of producing a good
translation, is significantly {\it improved} by pruning, and is not
degraded by grammar specialization. (The slight improvement in
coverage with EBL on is not statistically significant). Our
interpretation of these results is that the technical loss of grammar
coverage due to the specialization and pruning processes is more than
counterbalanced by two positive effects. Firstly, fewer utterances
time out due to slow processing; secondly, the reduced space of
possible analyses means that the problem of selecting between
different possible analyses of a given utterance becomes easier.

To sum up, the methods presented here demonstrate that it is possible
to use the combined pruning and grammar specialization method to speed
up the whole analysis phase by nearly an order of magnitude, without
incurring any real penalty in the form of reduced coverage.  We find
this an exciting and significant result, and are further continuing
our research in this area during the coming year. In the last two
paragraphs we sketch some ongoing work.

All the results presented above pertain to English only. The first
topic we have been investigating is the application of the methods
described here to processing of other languages. Preliminary
experiments we have carried out on the Swedish version of the CLE
\cite{GambackRayner:92} have been encouraging; using exactly the same
pruning methods and EBL chunking criteria as for English, we obtain
comparable speed-ups. The loss of coverage due to grammar
specialization also appears comparable, though we have not yet had
time to do the work needed to verify this properly. We intend to do so
soon, and also to repeat the experiments on the French version of the
CLE \cite{RaynerEtAl:96}.

The second topic is a more radical departure, and can be viewed as an
attempt to make interleaving of parsing and pruning the basic
principle underlying the CLE's linguistic analysis process.
Exploiting the ``stratified'' nature of the EBL-specialized grammar,
we group the chunked rules by level, and apply them one level at a
time, starting at the bottom. After each level, constituent pruning is
used to eliminate unlikely constituents. The intent is to achieve a
trainable robust parsing model, which can return a useful partial
analysis when no single global analysis is found.  An initial
implementation exists, and is currently being tested; preliminary
results here are also very positive. We expect to be able to report on
this work more fully in the near future.

\section*{Acknowledgements}

The work reported in this paper was funded by Telia Research AB.
We would like to thank Christer Samuelsson for making the
LR compiler available to us, Martin Keegan for patiently
judging the results of processing 16,000 ATIS utterances,
and Steve Pulman and Christer Samuelsson for helpful comments.

\section*{Appendix: definition of the ``New'' chunking rules}

\setlength{\itemsep}{-0.1in}

This appendix defines the ``New'' chunking rules referred to in
Sections~\ref{Section:EBL} and~\ref{Experiments}. There are seven
types of non-phrasal constituent in the specialised grammar.
We start by describing each type of constituent through examples.
\begin{description}
\item[Utterance:] The top category.
\item[Utterance\_unit:] {\tt Utterance\_ unit}s are minimal syntactic
units capable of standing on their own: for example, declarative
clauses, questions, NPs and PPs. Utterances may consist of more than
one {\tt utterance\_unit}. The following is an {\tt utterance} containing
two {\tt utterance\_unit}s: ``[Flights to Boston on Monday] [please
show me the cheapest ones.]''
\item[Imperative\_VP:] Since imperative verb phrases are very common
in the corpus, we make them a category of their own in the specialised
grammar. To generalise over possible addition of adverbials (in
particular, ``please'' and ``now''), we define the {\tt
imperative\_vp} category so as to leave the adverbials outside. Thus
the bracketed portion of the following utterance is an {\tt
imperative\_vp}:
``That's fine now [give me the fares for those flights]''
\item[Non\_phrasal\_NP:] All NPs which are not produced entirely
by phrasal rules. The following are all {\tt non\_phrasal\_NP}s:
``Boston and Denver'',  
``Flights on Sunday morning'', 
``Cheapest fare from Boston to Denver'',
``The meal I'd get on that flight''
\item[Rel:] Relative clauses.
\item[VP\_modifier:] VPs appearing as NP postmodifiers. 
The bracketed portions of the following are {\tt VP\_modifiers}:
``Delta flights [arriving after seven P M]'' 
``All flights tomorrow [ordered by arrival time]''
\item[PP:] The CLE grammar treats nominal temporal adverbials,
sequences of PPs, and ``A to B'' constructions as PPs (cf
\cite{SLT-report-grammar}). The following are examples of PPs:
``Tomorrow afternoon'', 
``From Boston to Dallas on Friday'', 
``Denver to San Francisco Sunday''
\end{description}
We can now present the precise criteria which determine the chunks of
rules composed to form each type of constituent. For each type of
constituent in the specialised grammar, the chunk is a subtree
extracted from the derivation tree of a training example (cf
\cite{SLT-report-EBL}); we specify the roots and leaves of the
relevant subtrees. The term ``phrasal tree'' will be used to mean a
derivation tree all of whose rule-applications are phrasal rules.
\begin{description}
\item[Utterance:] The root of the chunk is the root of the original
tree. The leaves are the nodes resulting from cutting at maximal
subtrees for {\tt utterance\_unit}s, {\tt non\_phrasal\_np}s {\tt
pp}s, and maximal phrasal subtrees.
\item[Utterance\_unit:] The root is the root of a maximal subtree for
a constituent of type {\tt utterance\_unit}. The leaves are the nodes
resulting from cutting at maximal subtrees for {\tt imperative\_vp}s, 
{\tt np}s, and {\tt pp}s, and maximal phrasal subtrees.
\item[Imperative\_VP:] The root is the root of a maximal subtree 
under an application of the {\tt S $\rightarrow$ VP} rule
whose root is not an application of an adverbial modification rule.
The leaves are the nodes
resulting from cutting at maximal subtrees for
{\tt non\_phrasal\_np}, and {\tt pp}, and maximal phrasal subtrees.
\item[Non\_phrasal\_NP:] The root is the root of a maximal non-phrasal
subtree for a constituent of type {\tt np}. The leaves are the nodes
resulting from cutting at maximal subtrees for  {\tt
rel}, {\tt vp\_modifier}, and {\tt pp}, and maximal phrasal subtrees.
\item[Rel:] The root is the root of a maximal subtree for
a constituent of type {\tt rel}. The leaves are the nodes resulting
from cutting at maximal subtrees for {\tt pp}, and maximal phrasal
subtrees.
\item[VP\_modifier:] The root is the root of a {\tt vp} subtree
immediately dominated by an application of the {\tt NP $\rightarrow$
NP VP} rule. The leaves are the nodes resulting from cutting at
maximal subtrees for {\tt pp}, and maximal phrasal subtrees.
\item[PP:] The root is the root of a maximal non-phrasal subtree for a
constituent of type {\tt pp}. The leaves are the nodes resulting from
cutting at maximal phrasal subtrees.
\end{description}

\end{document}